\documentclass{ceurart}

\usepackage{listings}
\usepackage{latexsym}
\usepackage{graphicx}
\usepackage{array}
\usepackage{booktabs}
\usepackage{caption}
\usepackage{lipsum}
\usepackage{tcolorbox}

\usepackage[T1]{fontenc}
\usepackage[utf8]{inputenc}
\usepackage{microtype}
\usepackage{inconsolata}
\tcbuselibrary{breakable}
\newtcolorbox{mybox}[1][]{
    colback=white,
    colframe=black,
    fonttitle=\bfseries,
    title=#1,
    width=\dimexpr\columnwidth-1em\relax,
    boxrule=0.5mm,
    colbacktitle=gray!20,
    coltitle=black,
    sharp corners,
    before=\par\smallskip\noindent,
    after=\par\smallskip,
    breakable
}

\newtcolorbox{mywhitebox}[1][]{
    colback=white
}

\begin{document}

\copyrightyear{2024}
\copyrightclause{Copyright for this paper by its authors.
  Use permitted under Creative Commons License Attribution 4.0
  International (CC BY 4.0).}

\conference{CHR 2024: Computational Humanities Research Conference, December 4–6, 2024, Aarhus, Denmark}

\title{Extracting Social Connections from Finnish Karelian Refugee Interviews Using LLMs}




\author[1]{Joonatan Laato}[
email=joonatan.m.laato@utu.fi,
]
\cormark[1]
\author[1]{Jenna Kanerva}[
email=jmnybl@utu.fi,
orcid=0000-0003-4580-5366,
]
\cormark[1]
\author[2]{John Loehr}[
email=john.loehr@helsinki.fi,
orcid=0000-0002-6212-0273,
]
\author[3]{Virpi Lummaa}[
email=virpi.lummaa@utu.fi,
orcid=0000-0002-2128-7587,
]
\author[1]{Filip Ginter}[
email=figint@utu.fi,
orcid=0000-0002-5484-6103,
]

\address[1]{TurkuNLP, Department of Computing, University of Turku, Finland}
\address[2]{Lammi Biological Station, Faculty of Biological and Environmental Sciences, University of Helsinki, Finland}
\address[3]{Department of Biology, University of Turku, Finland}

\cortext[1]{Equal contribution}

\begin{abstract}
We performed a zero-shot information extraction study on a historical collection of 89,339 brief Finnish-language interviews of refugee families relocated post-WWII from Finnish Eastern Karelia. Our research objective is two-fold. First, we aim to extract social organizations and hobbies from the free text of the interviews, separately for each family member. These can act as a proxy variable indicating the degree of social integration of refugees in their new environment. Second, we aim to evaluate several alternative ways to approach this task, comparing a number of generative models and a supervised learning approach, to gain a broader insight into the relative merits of these different approaches and their applicability in similar studies. 

We find that the best generative model (GPT-4) is roughly on par with human performance, at an F-score of 88.8\%. Interestingly, the best open generative model (Llama-3-70B-Instruct) reaches almost the same performance, at 87.7\% F-score, demonstrating that open models are becoming a viable alternative for some practical tasks even on non-English data. Additionally, we test a supervised learning alternative, where we fine-tune a Finnish BERT model (FinBERT) using GPT-4 generated training data. By this method, we achieved an F-score of 84.1\% already with 6K interviews up to an F-score of 86.3\% with 30k interviews. Such an approach would be particularly appealing in cases where the computational resources are limited, or there is a substantial mass of data to process.
\end{abstract}


\maketitle

\section{Introduction}

Numerous research questions in digital humanities require, in one form or another, extraction of factual information from very large textual collections \citep{ehrmann2023named}, often taking into account context spanning beyond the immediate textual neighbourhood. Traditionally, such tasks have been approached by first annotating a sufficiently large set of training instances, and subsequently building a dedicated supervised machine learning model for each task separately. Typically, these would be document- or entity-level annotations, such as document labels, named entity categories, and pairwise typed relations between the extracted entities.

Recently, the zero-shot capabilities of large language models (LLMs) combined with decreasing inference costs, offer an attractive alternative path, whereby the same outcome is achieved by prompting a suitable generative LLM, with only a small number of instances informing the model of the task (referred to as zero-shot or in-context learning). However, it is still very much an open question, which tasks can be approached in this manner, what levels of accuracy can be expected, which models are suitable, and what limitations one should be aware of when applying such methodology.

In this paper, we aim to add to this understanding, pursuing a case study on a large collection of short, Finnish-language family history interviews of post-WWII refugee families from the Finnish Karelia region. We address a context-dependent information extraction task on this dataset, intended to facilitate history research on the degree of social integration of refugee families in their new environment. In this paper, we nevertheless focus on the methodological rather than historical aspects of the work, aiming to answer the following: 

\begin{enumerate}
    \item Can LLM prompting process historical non-English data at a useful level of accuracy?
    \item Are there substantial differences in the performance of various LLMs for the task?
    \item Is it feasible to use LLMs to produce training data for a dedicated lightweight model?
\end{enumerate}

We evaluate OpenAI's proprietary models, GPT-3.5 \cite{brown2020language} and GPT-4 \cite{achiam2023gpt}, and contrast their performance with various open-source models. Additionally, we compare the LLM performance to a much smaller and lightweight tagger model fine-tuned for the task using GPT-4 generated training data. Such an approach becomes particularly appealing in cases where running a generative language model for the full dataset would not be economical. The models are evaluated on a manually annotated sample, which we also use to estimate human performance on the task.

\section{Related Work}

The recent conversational LLMs have been applied to various NLP tasks in a zero-shot setting, where the task instructions are given directly in the model prompt. While many such studies report encouraging performance (e.g.\ reasoning and dialogue \citep{qin-etal-2023-chatgpt, abaskohi-etal-2024-benchmarking-large}, text classification \citep{tornberg2023chatgpt4, gilardi2023chatgpt, debess-etal-2024-good-bad}, and relation classification \citep{aguda-etal-2024-large-language}), also negative results have been reported \citep{edwards-camacho-collados-2024-language-models, koneru-etal-2024-large-language}, especially when compared against task-specific supervised models.

For example, in the area of named entity recognition (NER) and relation extraction (RE), both \citet{wei2024zeroshot} and \citet{qin-etal-2023-chatgpt} report the generative LLMs substantially lagging behind the state-of-the-art supervised models in both tasks when evaluated on English. \citet{abaskohi-etal-2024-benchmarking-large} report similar results when applying ChatGPT to Persian named entity recognition. While their NER results clearly lag behind supervised models, their experiments show the LLMs being highly competitive in some Persian tasks, indicating the model works well also on a language other than English. This is also supported by \citet{debess-etal-2024-good-bad}, who report GPT-4 performing remarkably well when compared to human annotators on Faroese sentiment analysis, as well as \citet{tarkka-etal-2024-automated} on Finnish emotion classification.

Given the previous work, our study focuses on a task close to named entity recognition, however, in contrast to works by \citet{wei2024zeroshot,qin-etal-2023-chatgpt,abaskohi-etal-2024-benchmarking-large}, we do not attempt to create a universal method capable of returning any given named entities from any given input text. Rather we focus only on two types of entities (namely social organizations and hobbies) from a specific text collection (Karelian refugee interviews). Therefore, we can design the prompt specifically for the targeted entity types and text collection.

Previously, a targeted information retrieval study has been conducted e.g.\ in \citet{polak2023extracting}, where a target-specific, multi-step pipeline was developed for extracting materials properties from English materials science articles, reaching close to 90\% F-score using GPT-4. However, the text domain (English scientific articles compared to Finnish historical interviews), as well as targeted entity types (materials properties compared to persons' social organizations and hobbies), greatly differ in our study.

\section{Data}

\begin{figure*}[]
  \centering
  \includegraphics[width=\textwidth]{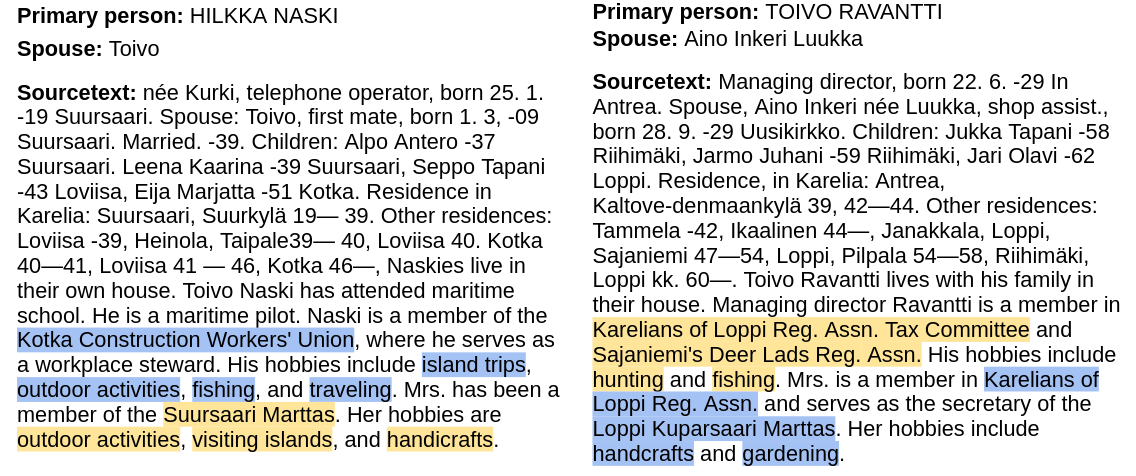}
  \caption{Two interviews translated into English for illustration purposes. The relevant information (hobbies, social organizations) related to the primary person (the one being interviewed) is illustrated in yellow, while blue indicates the spouse.}
  \label{fig:sample_stories}
\end{figure*}

Our primary source of data is \emph{Siirtokarjalaisten tie (The Journey of Displaced Karelians)} \citep{siirtokarjalaisetbook}, a 4-volume book series recording personal information and migration stories of Finnish citizens permanently displaced from the Eastern Karelia territories that Finland lost as the result of the 1939--1944 war with USSR. In total, there were around 420,000 refugees of which approximately 160,000 adults are represented in the interviews. The volumes consist of 89,339 family interviews conducted in 1968--1970 by approximately 300 trained interviewers. For each family, the interview covers their personal information, residential history in Karelia, and subsequent settlements across Finland. The interviews are semi-structured, where the basic demographic data (such as name, birth date and place, profession, information about spouse and children, and the family's residential history) is presented in a standardized format. However, the interviews contain also a less structured section including additional information about the family. This free-text part adds details about the way of accommodation (farm, house, apartment), size of the farm/property, wartime occupation and merits, as well as current social organizations and hobbies, among others.

In this paper, we build upon the work of \citet{loehr2017newly}, who produced a digitized version of the original books by scanning the pages and running an optical character recognition (OCR) on them. Additionally, they extracted a great amount of standardized demographic data using text patterns. Our primary focus is on extracting from the free-text interview information about individuals' reported memberships in different social organizations, as well as their hobbies and other free-time activities. Due to varying factual information present in the free-text part of the interviews, varying order of presentation, and non-standardized writing and vocabulary, such information cannot be easily extracted based on predefined text patterns. The second complicating factor is that in addition to extracting the relevant textual entities (such as \emph{hunting club}, or \emph{fishing}), we also need to associate these with the appropriate individuals, namely the husband or wife of the family, which is not always explicitly mentioned. Therefore, understanding the content of the interview as a whole is crucial to accurately determine the allocation of hobbies and organizations to the correct individuals. Two example interviews are shown in Figure~\ref{fig:sample_stories}, where the social organizations and hobbies associated with the primary person (the one being interviewed) are highlighted in yellow, and those associated with their spouse are illustrated in blue.

\subsection{Manually Annotated Evaluation Sample}

To measure the performance of different methods on the extraction task, we first created a manually annotated evaluation set. We randomly sampled 400 interviews and manually extracted the relevant information from them, dividing the information into four categories: \emph{primary person hobbies}, \emph{primary person social organizations}, \emph{spouse hobbies}, and \emph{spouse social organizations}. Professions, work titles, and wartime occupations were explicitly excluded from the annotation, as the study concentrates on free time activities. However, we decided to include some work-related actions as well, e.g.\ \emph{trade union}, \emph{workplace steward} or \emph{lay judge}, if such were considered to be additional, voluntary activities on top of the actual occupation indicating increased social connections.

As the reporting often included relatively free wording, we opted for doing the annotation by manually listing the extracted entities as plain text rather than doing strict span-level annotation specifying their exact location in the original reports. This approach allowed a more flexible interpretation of cases where the hobbies and organizations were not explicitly mentioned as single entities, however, we followed the original text as strictly as possible and avoided unnecessary rewriting. Nevertheless, considering the inflectional nature of the Finnish language as well as occasional errors originating from the OCR process, the extracted entities were normalized to their base forms, and the OCR-related errors were corrected (mostly including incorrect punctuation or spacing).

Additionally, if the person's title in the organization (such as \emph{member} or \emph{chairman}) was mentioned together with the organization name, the title was included in the annotation. However, if the title was not in direct connection with the organization name (which happened e.g.\ with complicated coordinations), we excluded it. In cases where a social organization was explicitly mentioned as a hobby, we preferred to annotate those as organizations (e.g.\ \emph{Her hobby is singing in the choir.}).

With this annotation approach, the manually annotated sample resembles the expected output of generative language models, which also give the extracted entities as plain text possibly including some amount of normalization. Using a similar approach also in manual annotation makes the language model evaluation more straightforward. In respect of the broader context of the study, the extraction of the interesting entities, in general, is considered more relevant than their exact wording, supporting also the decision of not focusing on exact spans.

The evaluation sample was fully double annotated, where two annotators independently went through each interview and extracted the relevant information. These individual annotations were then merged, and all disagreements were discussed and resolved together. Therefore, these merged consensus annotations constitute the final evaluation sample.

\subsection{Evaluation and Annotation Agreement}

The manually annotated evaluation sample contains 400 interviews, including a total of 1,478 annotated entities, which are divided quite evenly between hobbies and social organizations (50.8\% and 49.2\% respectively). 64.1\% of the entities belong to the primary person mentioned in the interview, while 35.9\% belong to the spouse. The sample is divided 50/50 into development and test sections, each containing 200 interviews. While the development section is used to optimize the prompt and parameters used in modelling, the test section is preserved for the final evaluation only.


To measure the reliability of the manually annotated gold standard, we approach the annotator agreement from two perspectives. First, we measure the inter-annotator agreement by comparing the two individual annotations against each other (without utilizing the final consensus annotations). The inter-annotator-agreement is calculated in terms of F-score, which is shown to approximate the traditional chance corrected agreement measures (like Cohen's kappa) in cases where the number of true negatives is large, which often is the case in information retrieval style of problems \citep{hripcsak2005agreement}. 

While precision, recall and F-score could be calculated straightforwardly by comparing the extracted entities of different annotators (producing exact match F-score), such a strict approach would not take into account minor variations in the extracted entities (such as differences in normalization). Therefore, to compare two entities we utilize fuzzy string matching using Levenshtein's normalized indel similarity (minimum number of character insertions and deletions required to transform the string into another) and set a threshold to determine the allowable similarity for two strings to be considered a match. By setting the threshold too high, we would miss cases where the retrieved entity is correct but the wording differs too much (e.g., \emph{Martha Organization / Marthas}) while setting the threshold too low would cause the evaluation to incorrectly pair similarly looking but different entities (e.g., \emph{theatre committee  / cultural committee}). By empirically inspecting the development section results, we set the threshold to 0.75 (1.0 indicating an exact match, while 0.0 would result in any two strings being recognized as a match). Setting the threshold any lower would run the risk of considering a reference entity correctly retrieved, while it was merely incorrectly paired due to the fuzzy matching threshold being too loose.

By using the micro F-score with the above-mentioned fuzzy matching threshold, the agreement of the two annotators is 83.4\% when comparing the two individual annotations against each other on the test section annotations. Secondly, we measure the two annotators against the final, consensus sample using the same evaluation metric. Note that when evaluating this way, the outcome is directly comparable to the evaluation later used in the language model experiments as well, giving us estimates of human performance on the task. When measuring against the final, consensus sample, the two annotators receive micro F-scores of 96.9\% and 85.6\% on the test section. 



In addition to the F-score, we separately measure how many of the fields (\emph{primary person hobbies}, \emph{primary person social organizations}, \emph{spouse hobbies}, \emph{spouse social organizations}) left empty in the gold annotation were correctly left empty also by the annotators. Both annotators have an accuracy above 99\% in these cases, indicating it was quite easy to notice the cases where there were no mentions of a specific property.

















\section{Data Extraction with LLMs}

\subsection{Preliminary Prompt Engineering}

We carried out a series of preliminary experiments with different prompts and a variety of different interviews in the ChatGPT web interface. During the prompt engineering phase, we experimented with both GPT-3.5 and GPT-4. From the beginning, it was clear that there was a substantial difference between these models (GPT-4 seeming far superior for the task). 

Our initial prompt started concise and simple: \emph{"From the following interview, extract hobbies and social organizations."} Although the model quite reliably succeeded with the initial prompt as well, we got some amount of edge cases where the model would do something unexpected, and not follow the given output format. Even if some of these happened quite infrequently, the same patterns occurred repeatedly when running a larger number of interviews through the model. For instance, the model suggested creating a Python script for the extraction rather than responding with the extracted data itself, explained its behaviour verbosely, or translated the findings into English. Therefore, we added to the prompt a set of explicit instructions to avoid such behaviour.

Another challenge was distinguishing information associated with the husband versus the wife. While the primary person (i.e.\ the one being interviewed) can be either one, the husband's activities are usually mentioned first. To handle this, we added to the prompt information about the general structure of the interviews and the typical order. Moreover, the LLMs were inconsistent in whether they gave the answers in a normalized form (e.g.\ removing inflection) or in the exact form as they appeared in the original text. The final prompt addressing all the issues above is shown in Figure \ref{fig:example_prompt}.

\begin{figure}[]
\centering
\begin{minipage}{\columnwidth}
\begin{mywhitebox}[]
\footnotesize 
\begin{verbatim}
I need you to scrape data from this text. 
These are interviews from Karelian people written in Finnish. 
List me names, IDs, hobbies, and social organizations. 
Notice to list spouse's hobbies and social orgs separately. 
Keep in mind that the husband is usually listed first in the story, even when he 
is not the primary person. 

Do not list jobs or wartime occupations. 
Do not suggest making an algorithm. 
Do not list the source text. 
Do not answer anything but the asked information. 
Do not translate your findings. Give your answers in base form. 
Differentiate one hobby or organization with a comma. 
Always respond in the following format for each story:

PersonName:
PersonID:
PersonHobbies:
PersonSocialOrgs:
SpouseName:
SpouseID:
SpouseHobbies:
SpouseSocialOrgs:

-----

primary_person_name: TOIVO JOHANNES JANATUINEN
primary_person_id: siirtokarjalaiset_2_15772P
spouse_name: Hanna Pukarinen
spouse_id: siirtokarjalaiset_2_15772S_1
source_text: <<interview text>>

\end{verbatim}
\end{mywhitebox}
\end{minipage}
\caption{The final prompt as used in our experiments paired with a sample interview.}
\label{fig:example_prompt}
\end{figure}


\begin{figure}[h!]
\centering
\begin{minipage}{\columnwidth}
\begin{mywhitebox}[]
\footnotesize 
\begin{verbatim}

PersonName: TOIVO RAVANTTI
PersonHobbies: hunting, fishing
PersonOrgs: Karelians of Loppi Reg. Assn., Sajaniemi's Deer Lads Reg. Assn.

SpouseName: Aino Inkeri Luukka
SpouseHobbies: handcrafts, gardening
SpouseOrgs: Karelians of Loppi Reg. Assn., Loppi's Kuparsaari Marttas
\end{verbatim}
\end{mywhitebox}
\end{minipage}
\caption{Example response obtained from the GPT-4 API (translated from Finnish).}
\label{fig:gpt4_response}
\end{figure}

\subsection{Processing LLM Response}

We used regular expressions to extract the key information from the language model output. An example LLM response is shown in Figure~\ref{fig:gpt4_response}. We identified the keywords, e.g.\ \texttt{PersonHobbies} or \texttt{SpouseSocialOrgs}, and extracted the model's responses following each keyword. This method allowed us to discard any unrelated details generated by the model. After the initial extraction, we further cleaned the data by removing frequently generated responses indicating no information was found, such as \emph{none}, \emph{N/A}, \emph{---}, or various language-specific phrases.


\begin{table*}
\caption{Feature ablation results on the development section. All open models use the English prompt with a single interview.}
\begin{tabular}{lllllll}
\hline
Run & Model & Pre & Rec & F-score \\\hline
\multicolumn{5}{l}{\textbf{GPT 3.5}}\\
\hspace{3mm}Prompt with single interview & gpt-3.5-turbo-1106 & 75.2 & 58.3 & 65.7   \\
\hspace{3mm}Prompt with two interviews & gpt-3.5-turbo-1106 & 75.8 & 51.8 & 61.6   \\
\hspace{3mm}Prompt with three interviews & gpt-3.5-turbo-1106 & 76.0 & 55.3 & 64.0   \\
\hspace{3mm}Finnish prompt with single interview & gpt-3.5-turbo-1106 & 66.4 & 58.8 & 62.3   \\
\multicolumn{5}{l}{\textbf{GPT 4}}\\
\hspace{3mm}Prompt with single interview & gpt-4-0613 & 89.9 & 85.3 & 87.5   \\
\hspace{3mm}Prompt with three interviews & gpt-4-0613 & 86.5 & 78.3 & 82.2   \\
\hspace{3mm}Finnish prompt with single interview & gpt-4-0613 & 91.2 & 88.5 & 89.8   \\
\hspace{3mm}Finnish prompt with single interview & gpt-4-Turbo-0125 & \textbf{92.0} & \textbf{88.8} & \textbf{90.4}   \\

\multicolumn{5}{l}{\textbf{Open models}}\\
\hspace{3mm}... & Llama-2-70B-Chat  & 45.3 & 49.4 & 47.3   \\
\hspace{3mm}... & Llama-3-70B-Instruct & \textbf{89.5} & \textbf{88.5} & \textbf{89.0}  \\
\hspace{3mm}... & Llama-3-8B-Instruct & 77.4 & 74.4 & 75.9  \\
\hspace{3mm}... & Mistral-7B-Instruct & 61.3 & 58.6 & 59.9   \\
\hspace{3mm}... & Mixtral-8x7B-Instruct & 77.7 & 74.6 & 76.2   \\
\hspace{3mm}... & Qwen-1.5-72B-Chat & 62.2 & 38.4 & 47.5   \\
\hspace{3mm}... & Qwen-2-72B-Instruct & 84.2 & 78.0 & 81.0   \\

\end{tabular}
\label{tab:devset-ablation-results}
\end{table*}



\subsection{Batching and Prompt Language}

Subsequently, we carried out a systematic evaluation of prompt language and batching using the development section of the manually annotated data using OpenAI's API interface.\footnote{\url{https://platform.openai.com/docs/api-reference/}} The numerical results are summarized in Table~\ref{tab:devset-ablation-results} and discussed below.

\paragraph{Batching} Our final prompt consists of 214 tokens, while the interviews average 328 tokens, varying between 50 and 800 tokens. Since the prompt length is significant compared to the average interview length and contributes to the inference cost, we evaluate whether multiple interviews can be batched into the same input. We experiment with batch sizes 1--3. However, as seen in Table~\ref{tab:devset-ablation-results}, for both GPT-3.5 and GPT-4, applying the prompt to each interview separately yields results several percentage points better in terms of F-score (+1.7pp for GPT-3.5 and +5.3pp for GPT-4 for 1 vs 3 interview batching). This result is consistent with prior work demonstrating that information extraction tends to be more effective at the beginning of a model's context window (e.g.\ \cite{kuratov2024search}).

\paragraph{Prompt language} As our source material is in Finnish, we compare whether the language of the prompt should match the language of the interviews, or remain in English. We noted a notable improvement of approximately +3pp with the GPT-4 when using a Finnish prompt. Interestingly for GPT-3.5, the results were clearly worse with a -3pp loss when prompted in Finnish.

\paragraph{} In summary, the best results using OpenAI's API were obtained with the newest GPT-4 model and a single interview Finnish prompt, obtaining an F-score of 90.4\%, while the GPT-3.5 model performed substantially worse.

\subsection{Open Models}

In addition to OpenAI's API-based models, we also experimented with various instruction fine-tuned, open models available through the HuggingFace infrastructure.\footnote{\url{https://huggingface.co/}} We test the following models:

\begin{description}
    
\item[Llama-2-70B-Chat] is a large conversational model released by Meta \cite{touvron2023llama}. The model is trained using mostly English data (90\%) while having only 0.03\% Finnish in its training data.

\item[Llama-3-8/70B-Instruct] is the latest model in Meta's Llama series \cite{llama3modelcard}, pretrained on over 15 trillion tokens from publicly available sources. While the model is primarily intended for English, it is also known to perform well in other languages, including Finnish.

\item[Mistral-7B-Instruct-v0.3] is the newest version of the smallest LLM by Mistral AI \cite{jiang2023mistral}. Mistral-7B supports officially only English and programming languages. 

\item[Mixtral-8x7B-Instruct-v0.1] is Mistral AI's Mixture of Experts (MoE) model \cite{jiang2024mixtral}. The model officially supports English, French, Italian, German, Spanish and programming languages, however, it was trained on multilingual data and shows some Finnish capability.

\item[Qwen-1.5-72B-Chat] is a conversational model released by Alibaba Cloud \cite{qwen}, focusing on English, Chinese, and programming languages, however, it was trained on multilingual data.

\item[Qwen-2-72B-Chat] is the latest development in the Qwen model series \cite{qwen}. 

\end{description}

For the open model experiments, we use the English language prompt, having noticed in preliminary runs that the open model outputs degraded when prompted in Finnish, some failing to even produce the correct format. A notable exception is Llama-3-70B-Instruct, only half a percentage point better on English prompt compared to Finnish. We also did not systematically test batching, since it was clear in our experiments with the commercial models that it would not lead to better results. 

The best results of all the open models were achieved by the latest version of the Llama series, Llama-3-70B-Instruct. This is a very encouraging result, as the performance of this open model is only marginally worse than the best commercial model (GPT-4). It is, however, important to note that 70B is a very large model and applying it on a non-trivial amount of input data inevitably requires access to supercomputing resources. 


The open models allow us to perform a hyperparameter optimization of the parameters that control text generation: greedy vs.\ sampled generation, and the sampling temperature for the latter. We carried out this optimization using Llama-3-70B-Instruct and found that lower temperatures produced better results for sampled generation (the best F-score of 88.1\% was reached with the temperature of 0.3), however, using a greedy generation without sampling yielded slightly better results, increasing the F-score to 89.0\%. This is perhaps unsurprising since the model is expected to rather straightforwardly extract information from the interviews, not to generate new output creatively.

We noticed that higher temperatures between 0.3-1.0 did encourage the model to answer more readily, improving recall at the cost of precision for an overall worse F-score. Temperature values above 1.0 already clearly degraded performance, and after a temperature value of 2.0, the model started to break format and hallucinate. We also experimented with generation beam search using widths up to 20, but this did not improve the performance.

\subsection{Error Analysis}

We conducted additional error analysis on the development set using the best-performing run of our top model, GPT-4, with Finnish prompt, i.e.\ the best setting as reported. Out of 809 entities (hobbies or organizations), the result comprised 58 false positives (FP), 82 false negatives (FN), and 669 true positives. We now proceed to analyse and categorize all errors in the predicted output.

\begin{description}

\item[Unmatched correct output (21 errors)] refers to entities that the model correctly extracted but were rephrased to such an extent that the fuzzy matching algorithm did not recognize them as matches using the 0.75 fuzzy match threshold (same as in the main experiments). Additionally, we manually verified that all fuzzy matches produced using this threshold were correct. Therefore, the fuzzy matching approach used in the automatic evaluation did not produce any incorrect pairings but did fail to recognize these 21 cases as matches.

\item[Straightforward omissions (37 errors)] were confirmed as cases where the model failed to identify the entities. Of these, 28 were social organizations and 9 were hobbies. In these cases, we could not find any indication in the output of the model as to why the entities were not extracted.

\item[Added or misinterpreted text (22 errors)] were mistakes where the model added or misinterpreted text, deviating from our original purpose and prompt. These primarily include extracting unprompted information such as competitions, awards, courses, and training (2 errors), occupations, side jobs, and wartime occupations (7 errors); as well as a number of varied unsystematic failures.

\item[Incorrect person attribution (17 errors)] are errors that can be attributed to the model misidentifying the person to which the entity attributes to (i.e.\ confusing the husband and the wife). Interestingly, though, these 17 errors were traced to a single, complex interview with a large number of entities. In general, and perhaps somewhat surprisingly, the model was -- barring this single interview -- flawless in inferring the person attribution. Note that the Finnish language does not mark gender in any manner, not even in personal pronouns (i.e.\ it lacks the distinction between \emph{he} and \emph{she}).

\item[Hobby vs organization (3 errors)] are cases where the model correctly extracted an entity but disagreed with the annotation on its categorization as a hobby or a social organization. These two categories do not in all cases have a crisp boundary, where, for example, \emph{theatre} in \emph{``He belonged to a theatre''} is annotated as an organization, but could plausibly be categorized also as a hobby.

\item[Entity boundary (7 errors)] are cases where the model deviated from the annotation in identifying boundaries of complex entities. For example, while the annotation identified \emph{verotarkastus, kansanhuolto- ja majoituslautakunta (tax inspection, public welfare, and accommodation committee)} as a single entity, the model divided it into three entities.

\end{description}

In summary, taking into account the evaluation errors produced by the fuzzy matching algorithm (counting also the 21 unmatched errors as correct predictions), we obtain precision \textbf{94.9}, recall \textbf{91.9}, and F-score \textbf{93.3} (compared to F-score of 90.4 calculated using our automatic evaluation). Note that the only difference here is the treatment of cases, where the automatic evaluation failed to match model predictions to their corresponding annotated counterparts. Since the manual evaluation also verified all automatically matched predictions being correct, the F-score 93.3 represents the true model performance on the development set, the automatic evaluation being slightly conservative and underestimating the real performance (as was intended). 

\subsection{Final Results on the Test Set}

\begin{table}[]
\caption{Test set numbers for different language models. All models are run with the best setting found during the development set experiments, therefore GPT-4-Turbo uses the Finnish single interview prompt, while the rest uses the English single interview prompt.}
\begin{tabular}{lllllll}
\hline
Model & Pre & Rec & F-score \\\hline
\hspace{3mm} GPT 3.5 & 63.8 & 63.1 & 63.4   \\
\hspace{3mm} GPT 4-Turbo & \textbf{90.5} & \textbf{87.2} & \textbf{88.8}  \\
\hspace{3mm} Llama-2-70B-Chat  & 55.3 & 46.6 & 50.6   \\
\hspace{3mm} Llama-3-70B-Instruct  & \textbf{87.9} & \textbf{87.5} & \textbf{87.7}   \\
\hspace{3mm} Llama-3-8B-Instruct & 75.9 & 75.4 & 75.6  \\
\hspace{3mm} Mixtral 8x7B-Instruct & 72.2 & 72.0 & 72.0  \\
\hspace{3mm} Mistral 7B-Instruct & 58.2 & 53.4 & 55.7  \\
\hspace{3mm} Qwen-1.5-72B-Chat & 67.4 & 40.7 & 50.7   \\
\hspace{3mm} Qwen-2-72B-Instruct & 83.4 & 78.0 & 80.6   \\
\end{tabular}
\label{tab:test-set-results}
\end{table}


Finally, all models were evaluated on the test section of the annotated data, with the best setting for each model as found during the evaluations on the development data. The results are summarized in Table~\ref{tab:test-set-results} and agree with the development set evaluations. The best LLM for the task, at F-score of 88.8\% is the commercial GPT-4 model. However, it is highly encouraging that the best open model Llama-3-70B-Instruct is only marginally worse at F-score of 87.7\%. All other tested models, commercial or open, perform worse by a large margin, with the order of the models being consistent with development set results except for Mixtral and Llama-3-8B, which change order in middle ranks. 

We measured GPU power consumption for small and large Llama-3 models running the 200 interview test set. The 8 billion parameter model consumed 119.63 Wh on a single AMD MI250x GPU. Estimated consumption for entire 89,339 interview extraction process with 8B model would add up to 53.48 kWh (without any efficiency optimization). In contrast power consumption with 70B model and 6 GPUs was 2609.06 Wh for 200 interviews. An estimated consumption for the entire run would add up to 1,165.49 kWh (without any efficiency optimization). All these calculations were done on CSC’s Lumi supercomputer, which runs on renewable hydro power.

\subsection{Full Extraction}

We processed all data using the GPT-4-Turbo-0125 model, extracting in total 193,194 hobbies and 161,108 organizations. Of the 89,339 total interviews, 79,594 had at least one identified entity. The full run costed a little over 1,000e when carried out, and at the time of writing the cost would be about 25\% of that. With our API response time limits, the full run took roughly 8 days, for an average of 8s per interview.


For 1.3\% of interviews, the model failed to follow the expected output format. The vast majority of these failed cases were corrected by re-running the interview again, and the remaining cases which were systematically failing were corrected by switching the prompt to English. This indicates a potentially viable strategy whereby several prompts for the same task are created and applied in succession to resolve cases where, for one reason or another, the model systematically produces obviously incorrect output with the default prompt. For instance, testing to see if the model's responses are in the correct format can indicate whether it has properly interpreted the prompt. 

\section{Data Extraction with a Fine-Tuned Model}

While LLMs have proven to be highly suitable for our task, and even the commercial variants can process substantial amounts of data at a reasonable cost, it is clear that if our data was, say, 100x larger, LLMs would not be an option due to their prohibitive computational requirements. In such cases, a less demanding model would be needed, but as our evaluation showed, smaller LLMs performed considerably worse. 

To complement our LLM-based results, we evaluate an alternative setting, whereby an LLM is used to create training data for training in a supervised manner a considerably smaller model with inference costs negligible to a large LLM. Before the realm of the LLMs, the standard method for our task would be to fine-tune a pre-trained encoder-only language model (BERT \citep{devlin-etal-2019-bert} or similar) as a token tagger. Modelling after the named entity recognition (NER) task, the set of entity classes would be organization and hobby, separated for the two persons in the interview, for a total of 4 entity classes.

For the experiments, we use the GPT-4 extracted entities as the training data for a standard implementation of a NER tagger based on the FinBERT model \cite{virtanen2019multilingual}, the de facto standard pre-trained encoder model for Finnish.

To create a dataset suitable for a NER model, it is necessary to match the extracted hobbies and organizations to their appropriate spans in the original text as well as give them the correct label, which is one of \texttt{P-HOB} and \texttt{P-ORG} for the primary person organization and hobby, and \texttt{S-HOB} and \texttt{S-ORG} for the spouse. We apply the standard IOB coding to account for multi-word entities.  

\begin{figure}[]
  \centering
  \includegraphics[width=1.0\textwidth]{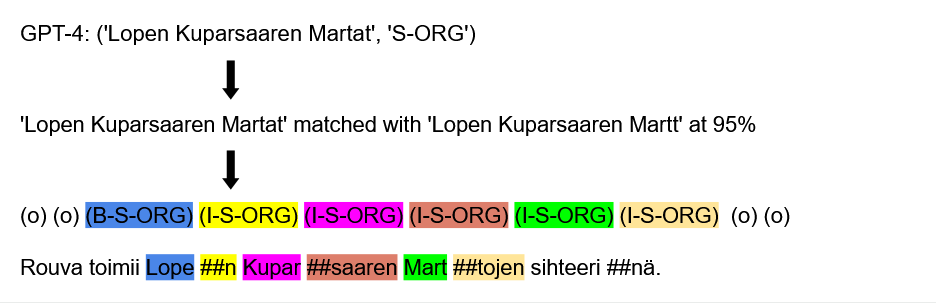}
  \caption{Process of creating NER-like data for model fine-tuning.
  English translation of the example is: \emph{Mrs.\ serves as the secretary of the Loppi Kuparsaari Martha's Association.} Hash symbols (\#\#) indicate the subword tokenization as produced by the FinBERT language model.}
  \label{fig:ner}
\end{figure}

In order to match the extracted entities in the original text, we aligned each entity with the interview text using the Levenshtein edit distance algorithm and considered all entities in their decreasing order by length to make sure the longest possible span is captured preferentially. We set a threshold of 60\% string similarity for all entity matches, calculated as edit distance divided by entity length in characters. If there was no match found for an entity, it was discarded. This process results in a dataset comprising of tokenized texts, and for each token, a label encoding the entity type to which the token belongs (including the relevant person in the interview), or the fact that the token does not belong to any entity. A single example is illustrated in Figure~\ref{fig:ner}.

\begin{figure}[]
  \centering
  \includegraphics[width=0.8\linewidth]{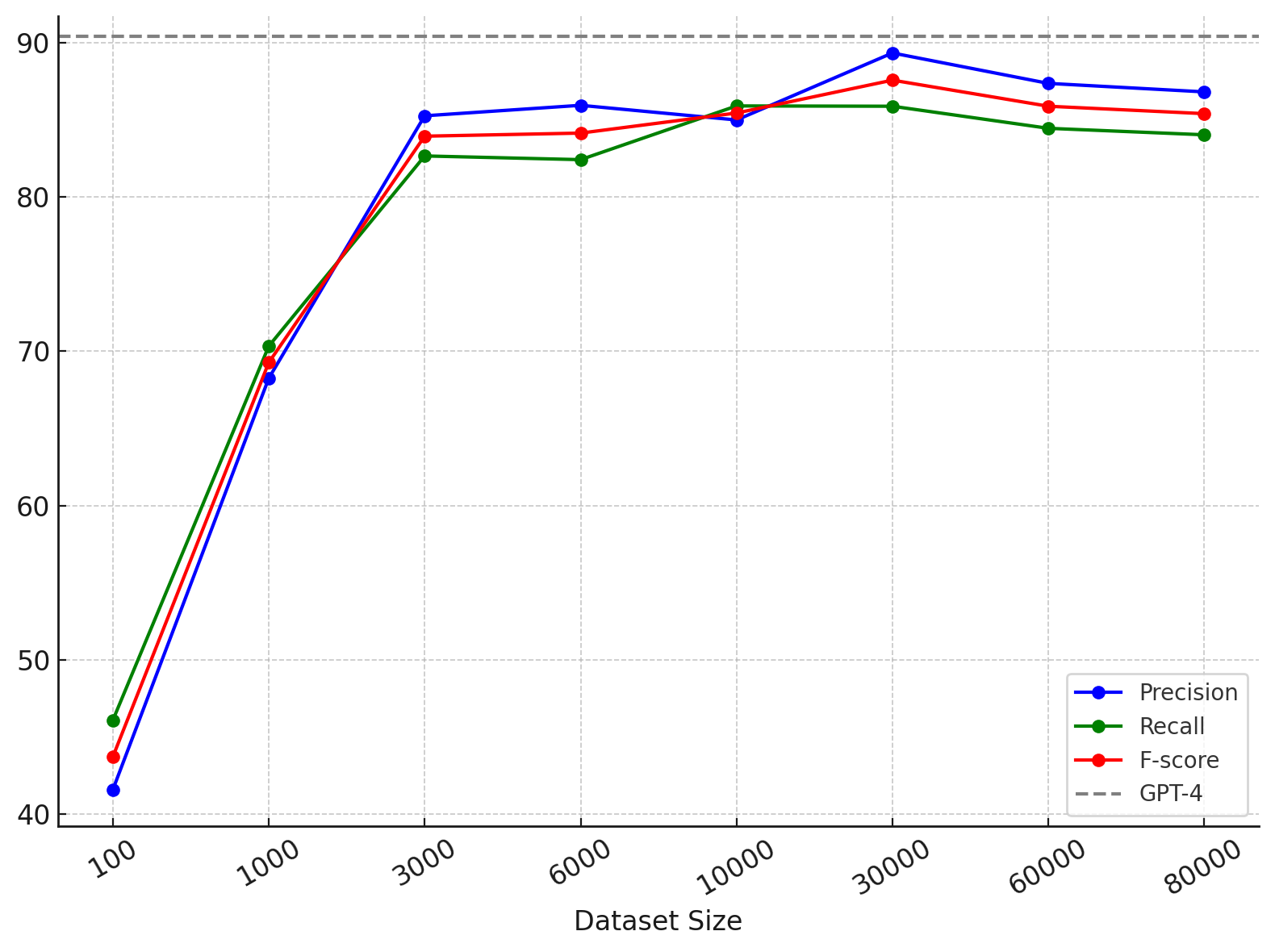}
  \caption{Performance of a fine-tuned BERT model with increasing training dataset size in terms of the number of training interviews. The model peaks at 30k training examples, corresponding of using 33\% of the full data for training (30k/90k). However, quite competitive results are already obtained when using 3k training examples (3\% of the full data). The dashed line represents the F-score of GPT-4.}
  \label{fig:ner_results}
\end{figure}

Our primary interest is to establish whether such a model can match the LLM results, and if so, what is the number of necessary training examples. To this end, we train a number of models with a progressively increasing training set size. Training hyperparameters are set on the development set and the results are measured on the test set, producing results comparable with the previously reported LLM results. The performance is shown in Figure~\ref{fig:ner_results}.



The tagger achieves a competitive performance after around 3,000 training interviews, at which point it outperforms all models except GPT-4 and Llama-3-70B-Instruct, the best LLMs in the previous experiments. The tagger continues to improve up to 30,000 training interviews and peaks at test set performance of precision 87.6\%, recall 85.0\%, and F-score of 86.3\%. This is only marginally below the performance of the considerably larger GPT-4 (F-score 88.8\%) and Llama-3-70B (F-score 87.7\%) models. 




\section{Discussion and Conclusions}

The Karelia refugee interview data consisted of 89,399 interviews of families, ranging in length between 50--800 tokens (about 20--500 words). The task was to retrieve hobbies and social organizations, separately for the husband and wife of each family. The task is of interest in its own right and also can serve as a model task of relatively straightforward information extraction that needs to be carried out on a very large textual dataset written in a small European language, and for which no specific pre-existing models are available.

Firstly, we find that the zero-shot capabilities of LLMs on this task approach human performance, as measured on a doubly annotated test dataset. Even for the best commercial model, the cost is acceptable, making it a viable option. We also establish that the best open model performs only marginally worse than the best commercial model. This is encouraging, as it demonstrates the closing gap between the commercial and open models in many practical applications, and opens up opportunities in cases where commercial API-based models are not applicable due to e.g.\ cost or limitations on data distribution. However, this holds only for the largest open models, which come with their own computational challenges. We also demonstrate that LLMs, whether commercial or open, are not a ``magic wand'' and that effort needs to be invested in evaluating different models and prompts on human-validated data.

Additionally, we report that, for our task and most likely for similar tasks as well, the approach of fine-tuning a lightweight encoder model using NER-like training data created by LLMs results in a very competitive performance. This would be a particularly suitable path to follow in cases where the amount of data to process would be orders of magnitude larger than our present case.

Our approaches benefited from the interviews being short enough to easily fit into the context length of present-day language models and also the format of the interviews being consistent enough that through rigorous prompt engineering, we were able to address most of the edge cases. Our conclusions are therefore most valid for tasks with similar properties.

Our future work will involve normalization and grounding of the extracted entities, for instance identifying that \emph{``Marttakerho (Martha club)''} and \emph{``Martat (Marthas)''} refer essentially to the same organization. We will also categorize the entities into a smaller number of more general categories (e.g.\ sports clubs or trade unions) enhancing the usability of the data for downstream analysis. 


Furthermore, future plans include combining the extracted data with information from other sources, such as family and pedigree information from digitized church books, and data on income, education, socio-economic group, cause of death etc.\ from the Population Register Centre of Finland and Statistics Finland. This unique dataset, drawing on a very substantial body of data, allows us to investigate how social and environmental contexts affect integration, social networks, reproduction, lifespan, and cause of death. With the extracted hobbies and social organizations, we aim to construct an estimate of individuals' social activity and identify aspects of sociality that have the greatest impact on migrants’ lives, a question highly relevant in today's world.

\section*{Acknowledgements}

This work was carried out in the \textit{Human Diversity} University profilation programme (PROFI-7) of the Research Council of Finland, and in part supported also through the \textit{Behind the Words} general research grant of the Research Council of Finland and the \textit{KinSocieties}, ERC-2022-ADG grant number 101098266. Computational resources were provided by CSC --- the Finnish IT Center for Science. 


\bibliography{main}

\begin{thebibliography}{25}
\expandafter\ifx\csname natexlab\endcsname\relax\def\natexlab#1{#1}\fi
\providecommand{\url}[1]{\texttt{#1}}
\providecommand{\href}[2]{#2}
\providecommand{\path}[1]{#1}
\providecommand{\DOIprefix}{doi:}
\providecommand{\ArXivprefix}{arXiv:}
\providecommand{\URLprefix}{URL: }
\providecommand{\Pubmedprefix}{pmid:}
\providecommand{\doi}[1]{\href{http://dx.doi.org/#1}{\path{#1}}}
\providecommand{\Pubmed}[1]{\href{pmid:#1}{\path{#1}}}
\providecommand{\bibinfo}[2]{#2}
\ifx\xfnm\relax \def\xfnm[#1]{\unskip,\space#1}\fi
\bibitem[{{Ehrmann} et~al.(2023){Ehrmann}, {Hamdi}, {Pontes}, {Romanello}, and {Doucet}}]{ehrmann2023named}
\bibinfo{author}{M.~{Ehrmann}}, \bibinfo{author}{A.~{Hamdi}}, \bibinfo{author}{E.~L. {Pontes}}, \bibinfo{author}{M.~{Romanello}}, \bibinfo{author}{A.~{Doucet}},
\newblock \bibinfo{title}{Named entity recognition and classification in historical documents: A survey},
\newblock \bibinfo{journal}{ACM Computing Surveys} \bibinfo{volume}{56} (\bibinfo{year}{2023}) \bibinfo{pages}{1--47}.
\bibitem[{{Brown} et~al.(2020){Brown}, {Mann}, {Ryder}, {Subbiah}, {Kaplan}, {Dhariwal}, {Neelakantan}, {Shyam}, {Sastry}, {Askell} et~al.}]{brown2020language}
\bibinfo{author}{T.~{Brown}}, \bibinfo{author}{B.~{Mann}}, \bibinfo{author}{N.~{Ryder}}, \bibinfo{author}{M.~{Subbiah}}, \bibinfo{author}{J.~D. {Kaplan}}, \bibinfo{author}{P.~{Dhariwal}}, \bibinfo{author}{A.~{Neelakantan}}, \bibinfo{author}{P.~{Shyam}}, \bibinfo{author}{G.~{Sastry}}, \bibinfo{author}{A.~{Askell}}, et~al.,
\newblock \bibinfo{title}{Language models are few-shot learners},
\newblock \bibinfo{journal}{Advances in Neural Information Processing Systems} \bibinfo{volume}{33} (\bibinfo{year}{2020}) \bibinfo{pages}{1877--1901}.
\bibitem[{{Achiam} et~al.(2023){Achiam}, {Adler}, {Agarwal}, {Ahmad}, {Akkaya}, {Aleman}, {Almeida}, {Altenschmidt}, {Altman}, {Anadkat} et~al.}]{achiam2023gpt}
\bibinfo{author}{J.~{Achiam}}, \bibinfo{author}{S.~{Adler}}, \bibinfo{author}{S.~{Agarwal}}, \bibinfo{author}{L.~{Ahmad}}, \bibinfo{author}{I.~{Akkaya}}, \bibinfo{author}{F.~L. {Aleman}}, \bibinfo{author}{D.~{Almeida}}, \bibinfo{author}{J.~{Altenschmidt}}, \bibinfo{author}{S.~{Altman}}, \bibinfo{author}{S.~{Anadkat}}, et~al.,
\newblock \bibinfo{title}{{GPT-4} technical report},
\newblock \bibinfo{journal}{arXiv preprint arXiv:2303.08774}  (\bibinfo{year}{2023}).
\bibitem[{{Qin} et~al.(2023){Qin}, {Zhang}, {Zhang}, {Chen}, {Yasunaga}, and {Yang}}]{qin-etal-2023-chatgpt}
\bibinfo{author}{C.~{Qin}}, \bibinfo{author}{A.~{Zhang}}, \bibinfo{author}{Z.~{Zhang}}, \bibinfo{author}{J.~{Chen}}, \bibinfo{author}{M.~{Yasunaga}}, \bibinfo{author}{D.~{Yang}},
\newblock \bibinfo{title}{Is {ChatGPT} a general-purpose natural language processing task solver?},
\newblock in: \bibinfo{editor}{H.~{Bouamor}}, \bibinfo{editor}{J.~{Pino}}, \bibinfo{editor}{K.~{Bali}} (Eds.), \bibinfo{booktitle}{Proceedings of the 2023 Conference on Empirical Methods in Natural Language Processing}, \bibinfo{publisher}{Association for Computational Linguistics}, \bibinfo{address}{Singapore}, \bibinfo{year}{2023}, pp. \bibinfo{pages}{1339--1384}. \URLprefix \url{https://aclanthology.org/2023.emnlp-main.85}. \DOIprefix\doi{10.18653/v1/2023.emnlp-main.85}.
\bibitem[{{Abaskohi} et~al.(2024){Abaskohi}, {Baruni}, {Masoudi}, {Abbasi}, {Babalou}, {Edalat}, {Kamahi}, {Mahdizadeh Sani}, {Naghavian}, {Namazifard}, {Sadeghi}, and {Yaghoobzadeh}}]{abaskohi-etal-2024-benchmarking-large}
\bibinfo{author}{A.~{Abaskohi}}, \bibinfo{author}{S.~{Baruni}}, \bibinfo{author}{M.~{Masoudi}}, \bibinfo{author}{N.~{Abbasi}}, \bibinfo{author}{M.~H. {Babalou}}, \bibinfo{author}{A.~{Edalat}}, \bibinfo{author}{S.~{Kamahi}}, \bibinfo{author}{S.~{Mahdizadeh Sani}}, \bibinfo{author}{N.~{Naghavian}}, \bibinfo{author}{D.~{Namazifard}}, \bibinfo{author}{P.~{Sadeghi}}, \bibinfo{author}{Y.~{Yaghoobzadeh}},
\newblock \bibinfo{title}{Benchmarking large language models for {Persian}: A preliminary study focusing on {ChatGPT}},
\newblock in: \bibinfo{editor}{N.~{Calzolari}}, \bibinfo{editor}{M.-Y. {Kan}}, \bibinfo{editor}{V.~{Hoste}}, \bibinfo{editor}{A.~{Lenci}}, \bibinfo{editor}{S.~{Sakti}}, \bibinfo{editor}{N.~{Xue}} (Eds.), \bibinfo{booktitle}{Proceedings of the 2024 Joint International Conference on Computational Linguistics, Language Resources and Evaluation (LREC-COLING 2024)}, \bibinfo{publisher}{ELRA and ICCL}, \bibinfo{address}{Torino, Italia}, \bibinfo{year}{2024}, pp. \bibinfo{pages}{2189--2203}. \URLprefix \url{https://aclanthology.org/2024.lrec-main.197}.
\bibitem[{{T{\"o}rnberg}(2023)}]{tornberg2023chatgpt4}
\bibinfo{author}{P.~{T{\"o}rnberg}}, \bibinfo{title}{{ChatGPT-4} outperforms experts and crowd workers in annotating political {Twitter} messages with zero-shot learning}, \bibinfo{year}{2023}. \href{http://arxiv.org/abs/2304.06588}{{\tt arXiv:2304.06588}}.
\bibitem[{{Gilardi} et~al.(2023){Gilardi}, {Alizadeh}, and {Kubli}}]{gilardi2023chatgpt}
\bibinfo{author}{F.~{Gilardi}}, \bibinfo{author}{M.~{Alizadeh}}, \bibinfo{author}{M.~{Kubli}},
\newblock \bibinfo{title}{{ChatGPT} outperforms crowd workers for text-annotation tasks},
\newblock \bibinfo{journal}{Proceedings of the National Academy of Sciences} \bibinfo{volume}{120} (\bibinfo{year}{2023}) \bibinfo{pages}{e2305016120}.
\bibitem[{{Debess} et~al.(2024){Debess}, {Simonsen}, and {Einarsson}}]{debess-etal-2024-good-bad}
\bibinfo{author}{I.~N. {Debess}}, \bibinfo{author}{A.~{Simonsen}}, \bibinfo{author}{H.~{Einarsson}},
\newblock \bibinfo{title}{Good or bad news? {E}xploring {GPT}-4 for sentiment analysis for {Faroese} on a public news corpora},
\newblock in: \bibinfo{editor}{N.~{Calzolari}}, \bibinfo{editor}{M.-Y. {Kan}}, \bibinfo{editor}{V.~{Hoste}}, \bibinfo{editor}{A.~{Lenci}}, \bibinfo{editor}{S.~{Sakti}}, \bibinfo{editor}{N.~{Xue}} (Eds.), \bibinfo{booktitle}{Proceedings of the 2024 Joint International Conference on Computational Linguistics, Language Resources and Evaluation (LREC-COLING 2024)}, \bibinfo{publisher}{ELRA and ICCL}, \bibinfo{address}{Torino, Italia}, \bibinfo{year}{2024}, pp. \bibinfo{pages}{7814--7824}. \URLprefix \url{https://aclanthology.org/2024.lrec-main.690}.
\bibitem[{{Aguda} et~al.(2024){Aguda}, {Siddagangappa}, {Kochkina}, {Kaur}, {Wang}, and {Smiley}}]{aguda-etal-2024-large-language}
\bibinfo{author}{T.~D. {Aguda}}, \bibinfo{author}{S.~{Siddagangappa}}, \bibinfo{author}{E.~{Kochkina}}, \bibinfo{author}{S.~{Kaur}}, \bibinfo{author}{D.~{Wang}}, \bibinfo{author}{C.~{Smiley}},
\newblock \bibinfo{title}{Large language models as financial data annotators: A study on effectiveness and efficiency},
\newblock in: \bibinfo{editor}{N.~{Calzolari}}, \bibinfo{editor}{M.-Y. {Kan}}, \bibinfo{editor}{V.~{Hoste}}, \bibinfo{editor}{A.~{Lenci}}, \bibinfo{editor}{S.~{Sakti}}, \bibinfo{editor}{N.~{Xue}} (Eds.), \bibinfo{booktitle}{Proceedings of the 2024 Joint International Conference on Computational Linguistics, Language Resources and Evaluation (LREC-COLING 2024)}, \bibinfo{publisher}{ELRA and ICCL}, \bibinfo{address}{Torino, Italia}, \bibinfo{year}{2024}, pp. \bibinfo{pages}{10124--10145}. \URLprefix \url{https://aclanthology.org/2024.lrec-main.885}.
\bibitem[{{Edwards} and {Camacho-Collados}(2024)}]{edwards-camacho-collados-2024-language-models}
\bibinfo{author}{A.~{Edwards}}, \bibinfo{author}{J.~{Camacho-Collados}},
\newblock \bibinfo{title}{Language models for text classification: Is in-context learning enough?},
\newblock in: \bibinfo{editor}{N.~{Calzolari}}, \bibinfo{editor}{M.-Y. {Kan}}, \bibinfo{editor}{V.~{Hoste}}, \bibinfo{editor}{A.~{Lenci}}, \bibinfo{editor}{S.~{Sakti}}, \bibinfo{editor}{N.~{Xue}} (Eds.), \bibinfo{booktitle}{Proceedings of the 2024 Joint International Conference on Computational Linguistics, Language Resources and Evaluation (LREC-COLING 2024)}, \bibinfo{publisher}{ELRA and ICCL}, \bibinfo{address}{Torino, Italia}, \bibinfo{year}{2024}, pp. \bibinfo{pages}{10058--10072}. \URLprefix \url{https://aclanthology.org/2024.lrec-main.879}.
\bibitem[{{Koneru} et~al.(2024){Koneru}, {Wu}, and {Rajtmajer}}]{koneru-etal-2024-large-language}
\bibinfo{author}{S.~{Koneru}}, \bibinfo{author}{J.~{Wu}}, \bibinfo{author}{S.~{Rajtmajer}},
\newblock \bibinfo{title}{Can large language models discern evidence for scientific hypotheses? {C}ase studies in the social sciences},
\newblock in: \bibinfo{editor}{N.~{Calzolari}}, \bibinfo{editor}{M.-Y. {Kan}}, \bibinfo{editor}{V.~{Hoste}}, \bibinfo{editor}{A.~{Lenci}}, \bibinfo{editor}{S.~{Sakti}}, \bibinfo{editor}{N.~{Xue}} (Eds.), \bibinfo{booktitle}{Proceedings of the 2024 Joint International Conference on Computational Linguistics, Language Resources and Evaluation (LREC-COLING 2024)}, \bibinfo{publisher}{ELRA and ICCL}, \bibinfo{address}{Torino, Italia}, \bibinfo{year}{2024}, pp. \bibinfo{pages}{2787--2797}. \URLprefix \url{https://aclanthology.org/2024.lrec-main.248}.
\bibitem[{{Wei} et~al.(2024){Wei}, {Cui}, {Cheng}, {Wang}, {Zhang}, {Huang}, {Xie}, {Xu}, {Chen}, {Zhang}, {Jiang}, and {Han}}]{wei2024zeroshot}
\bibinfo{author}{X.~{Wei}}, \bibinfo{author}{X.~{Cui}}, \bibinfo{author}{N.~{Cheng}}, \bibinfo{author}{X.~{Wang}}, \bibinfo{author}{X.~{Zhang}}, \bibinfo{author}{S.~{Huang}}, \bibinfo{author}{P.~{Xie}}, \bibinfo{author}{J.~{Xu}}, \bibinfo{author}{Y.~{Chen}}, \bibinfo{author}{M.~{Zhang}}, \bibinfo{author}{Y.~{Jiang}}, \bibinfo{author}{W.~{Han}}, \bibinfo{title}{{ChatIE}: Zero-shot information extraction via chatting with {ChatGPT}}, \bibinfo{year}{2024}. \href{http://arxiv.org/abs/2302.10205}{{\tt arXiv:2302.10205}}.
\bibitem[{{Tarkka} et~al.(2024){Tarkka}, {Koljonen}, {Korhonen}, {Laine}, {Martiskainen}, {Elo}, and {Laippala}}]{tarkka-etal-2024-automated}
\bibinfo{author}{O.~{Tarkka}}, \bibinfo{author}{J.~{Koljonen}}, \bibinfo{author}{M.~{Korhonen}}, \bibinfo{author}{J.~{Laine}}, \bibinfo{author}{K.~{Martiskainen}}, \bibinfo{author}{K.~{Elo}}, \bibinfo{author}{V.~{Laippala}},
\newblock \bibinfo{title}{Automated emotion annotation of {Finnish} parliamentary speeches using {GPT}-4},
\newblock in: \bibinfo{editor}{D.~{Fiser}}, \bibinfo{editor}{M.~{Eskevich}}, \bibinfo{editor}{D.~{Bordon}} (Eds.), \bibinfo{booktitle}{Proceedings of the IV Workshop on Creating, Analysing, and Increasing Accessibility of Parliamentary Corpora (ParlaCLARIN) @ LREC-COLING 2024}, \bibinfo{publisher}{ELRA and ICCL}, \bibinfo{address}{Torino, Italia}, \bibinfo{year}{2024}, pp. \bibinfo{pages}{70--76}. \URLprefix \url{https://aclanthology.org/2024.parlaclarin-1.11}.
\bibitem[{{Polak} and {Morgan}(2023)}]{polak2023extracting}
\bibinfo{author}{M.~P. {Polak}}, \bibinfo{author}{D.~{Morgan}}, \bibinfo{title}{Extracting accurate materials data from research papers with conversational language models and prompt engineering}, \bibinfo{year}{2023}. \href{http://arxiv.org/abs/2303.05352}{{\tt arXiv:2303.05352}}.
\bibitem[{Anon(1970)}]{siirtokarjalaisetbook}
\bibinfo{author}{Anon},
\newblock \bibinfo{title}{Siirtokarjalaisten tie},
\newblock \bibinfo{journal}{Turku, Finland: Nyky-Karjala}  (\bibinfo{year}{1970}).
\bibitem[{{Loehr} et~al.(2017){Loehr}, {Lynch}, {Mappes}, {Salmi}, {Pettay}, and {Lummaa}}]{loehr2017newly}
\bibinfo{author}{J.~A. {Loehr}}, \bibinfo{author}{R.~{Lynch}}, \bibinfo{author}{J.~{Mappes}}, \bibinfo{author}{T.~{Salmi}}, \bibinfo{author}{J.~{Pettay}}, \bibinfo{author}{V.~{Lummaa}},
\newblock \bibinfo{title}{Newly digitized database reveals the lives and families of forced migrants from {Finnish} {K}arelia},
\newblock \bibinfo{journal}{{Finnish} Yearbook of Population Research}  (\bibinfo{year}{2017}).
\bibitem[{{Hripcsak} and {Rothschild}(2005)}]{hripcsak2005agreement}
\bibinfo{author}{G.~{Hripcsak}}, \bibinfo{author}{A.~S. {Rothschild}},
\newblock \bibinfo{title}{Agreement, the f-measure, and reliability in information retrieval},
\newblock \bibinfo{journal}{Journal of the American Medical Informatics Association} \bibinfo{volume}{12} (\bibinfo{year}{2005}) \bibinfo{pages}{296--298}. \DOIprefix\doi{https://doi.org/10.1197/jamia.M1733}.
\bibitem[{{Kuratov} et~al.(2024){Kuratov}, {Bulatov}, {Anokhin}, {Sorokin}, {Sorokin}, and {Burtsev}}]{kuratov2024search}
\bibinfo{author}{Y.~{Kuratov}}, \bibinfo{author}{A.~{Bulatov}}, \bibinfo{author}{P.~{Anokhin}}, \bibinfo{author}{D.~{Sorokin}}, \bibinfo{author}{A.~{Sorokin}}, \bibinfo{author}{M.~{Burtsev}},
\newblock \bibinfo{title}{In search of needles in a 10m haystack: Recurrent memory finds what llms miss},
\newblock \bibinfo{journal}{arXiv preprint arXiv:2402.10790}  (\bibinfo{year}{2024}).
\bibitem[{{Touvron} et~al.(2023){Touvron}, {Martin}, {Stone}, {Albert}, {Almahairi}, {Babaei}, {Bashlykov}, {Batra}, {Bhargava}, {Bhosale} et~al.}]{touvron2023llama}
\bibinfo{author}{H.~{Touvron}}, \bibinfo{author}{L.~{Martin}}, \bibinfo{author}{K.~{Stone}}, \bibinfo{author}{P.~{Albert}}, \bibinfo{author}{A.~{Almahairi}}, \bibinfo{author}{Y.~{Babaei}}, \bibinfo{author}{N.~{Bashlykov}}, \bibinfo{author}{S.~{Batra}}, \bibinfo{author}{P.~{Bhargava}}, \bibinfo{author}{S.~{Bhosale}}, et~al.,
\newblock \bibinfo{title}{Llama 2: Open foundation and fine-tuned chat models},
\newblock \bibinfo{journal}{arXiv preprint arXiv:2307.09288}  (\bibinfo{year}{2023}).
\bibitem[{AI@Meta(2024)}]{llama3modelcard}
\bibinfo{author}{AI@Meta}, \bibinfo{title}{Llama 3 model card}, \bibinfo{year}{2024}. \URLprefix \url{https://github.com/meta-llama/llama3/blob/main/MODEL_CARD.md}.
\bibitem[{{Jiang} et~al.(2023){Jiang}, {Sablayrolles}, {Mensch}, {Bamford}, {Chaplot}, de~las {Casas}, {Bressand}, {Lengyel}, {Lample}, {Saulnier} et~al.}]{jiang2023mistral}
\bibinfo{author}{A.~Q. {Jiang}}, \bibinfo{author}{A.~{Sablayrolles}}, \bibinfo{author}{A.~{Mensch}}, \bibinfo{author}{C.~{Bamford}}, \bibinfo{author}{D.~S. {Chaplot}}, \bibinfo{author}{D.~de~las {Casas}}, \bibinfo{author}{F.~{Bressand}}, \bibinfo{author}{G.~{Lengyel}}, \bibinfo{author}{G.~{Lample}}, \bibinfo{author}{L.~{Saulnier}}, et~al.,
\newblock \bibinfo{title}{Mistral 7b},
\newblock \bibinfo{journal}{arXiv preprint arXiv:2310.06825}  (\bibinfo{year}{2023}).
\bibitem[{{Jiang} et~al.(2024){Jiang}, {Sablayrolles}, {Roux}, {Mensch}, {Savary}, {Bamford}, {Chaplot}, de~las {Casas}, {Hanna}, {Bressand} et~al.}]{jiang2024mixtral}
\bibinfo{author}{A.~Q. {Jiang}}, \bibinfo{author}{A.~{Sablayrolles}}, \bibinfo{author}{A.~{Roux}}, \bibinfo{author}{A.~{Mensch}}, \bibinfo{author}{B.~{Savary}}, \bibinfo{author}{C.~{Bamford}}, \bibinfo{author}{D.~S. {Chaplot}}, \bibinfo{author}{D.~de~las {Casas}}, \bibinfo{author}{E.~B. {Hanna}}, \bibinfo{author}{F.~{Bressand}}, et~al.,
\newblock \bibinfo{title}{Mixtral of experts},
\newblock \bibinfo{journal}{arXiv preprint arXiv:2401.04088}  (\bibinfo{year}{2024}).
\bibitem[{{Bai} et~al.(2023){Bai}, {Bai}, {Chu}, {Cui}, {Dang}, {Deng}, {Fan}, {Ge}, {Han}, {Huang}, {Hui}, {Ji}, {Li}, {Lin}, {Lin}, {Liu}, {Liu}, {Lu}, {Lu}, {Ma}, {Men}, {Ren}, {Ren}, {Tan}, {Tan}, {Tu}, {Wang}, {Wang}, {Wang}, {Wu}, {Xu}, {Xu}, {Yang}, {Yang}, {Yang}, {Yang}, {Yao}, {Yu}, {Yuan}, {Yuan}, {Zhang}, {Zhang}, {Zhang}, {Zhang}, {Zhou}, {Zhou}, {Zhou}, and {Zhu}}]{qwen}
\bibinfo{author}{J.~{Bai}}, \bibinfo{author}{S.~{Bai}}, \bibinfo{author}{Y.~{Chu}}, \bibinfo{author}{Z.~{Cui}}, \bibinfo{author}{K.~{Dang}}, \bibinfo{author}{X.~{Deng}}, \bibinfo{author}{Y.~{Fan}}, \bibinfo{author}{W.~{Ge}}, \bibinfo{author}{Y.~{Han}}, \bibinfo{author}{F.~{Huang}}, \bibinfo{author}{B.~{Hui}}, \bibinfo{author}{L.~{Ji}}, \bibinfo{author}{M.~{Li}}, \bibinfo{author}{J.~{Lin}}, \bibinfo{author}{R.~{Lin}}, \bibinfo{author}{D.~{Liu}}, \bibinfo{author}{G.~{Liu}}, \bibinfo{author}{C.~{Lu}}, \bibinfo{author}{K.~{Lu}}, \bibinfo{author}{J.~{Ma}}, \bibinfo{author}{R.~{Men}}, \bibinfo{author}{X.~{Ren}}, \bibinfo{author}{X.~{Ren}}, \bibinfo{author}{C.~{Tan}}, \bibinfo{author}{S.~{Tan}}, \bibinfo{author}{J.~{Tu}}, \bibinfo{author}{P.~{Wang}}, \bibinfo{author}{S.~{Wang}}, \bibinfo{author}{W.~{Wang}}, \bibinfo{author}{S.~{Wu}}, \bibinfo{author}{B.~{Xu}}, \bibinfo{author}{J.~{Xu}}, \bibinfo{author}{A.~{Yang}}, \bibinfo{author}{H.~{Yang}}, \bibinfo{author}{J.~{Yang}}, \bibinfo{author}{S.~{Yang}},
  \bibinfo{author}{Y.~{Yao}}, \bibinfo{author}{B.~{Yu}}, \bibinfo{author}{H.~{Yuan}}, \bibinfo{author}{Z.~{Yuan}}, \bibinfo{author}{J.~{Zhang}}, \bibinfo{author}{X.~{Zhang}}, \bibinfo{author}{Y.~{Zhang}}, \bibinfo{author}{Z.~{Zhang}}, \bibinfo{author}{C.~{Zhou}}, \bibinfo{author}{J.~{Zhou}}, \bibinfo{author}{X.~{Zhou}}, \bibinfo{author}{T.~{Zhu}},
\newblock \bibinfo{title}{Qwen technical report},
\newblock \bibinfo{journal}{arXiv preprint arXiv:2309.16609}  (\bibinfo{year}{2023}).
\bibitem[{{Devlin} et~al.(2019){Devlin}, {Chang}, {Lee}, and {Toutanova}}]{devlin-etal-2019-bert}
\bibinfo{author}{J.~{Devlin}}, \bibinfo{author}{M.-W. {Chang}}, \bibinfo{author}{K.~{Lee}}, \bibinfo{author}{K.~{Toutanova}},
\newblock \bibinfo{title}{{BERT}: Pre-training of deep bidirectional transformers for language understanding},
\newblock in: \bibinfo{editor}{J.~{Burstein}}, \bibinfo{editor}{C.~{Doran}}, \bibinfo{editor}{T.~{Solorio}} (Eds.), \bibinfo{booktitle}{Proceedings of the 2019 Conference of the North American Chapter of the Association for Computational Linguistics: Human Language Technologies, Volume 1 (Long and Short Papers)}, \bibinfo{publisher}{Association for Computational Linguistics}, \bibinfo{address}{Minneapolis, Minnesota}, \bibinfo{year}{2019}, pp. \bibinfo{pages}{4171--4186}. \URLprefix \url{https://aclanthology.org/N19-1423}. \DOIprefix\doi{10.18653/v1/N19-1423}.
\bibitem[{{Virtanen} et~al.(2019){Virtanen}, {Kanerva}, {Ilo}, {Luoma}, {Luotolahti}, {Salakoski}, {Ginter}, and {Pyysalo}}]{virtanen2019multilingual}
\bibinfo{author}{A.~{Virtanen}}, \bibinfo{author}{J.~{Kanerva}}, \bibinfo{author}{R.~{Ilo}}, \bibinfo{author}{J.~{Luoma}}, \bibinfo{author}{J.~{Luotolahti}}, \bibinfo{author}{T.~{Salakoski}}, \bibinfo{author}{F.~{Ginter}}, \bibinfo{author}{S.~{Pyysalo}}, \bibinfo{title}{Multilingual is not enough: {BERT} for {Finnish}}, \bibinfo{year}{2019}. \href{http://arxiv.org/abs/1912.07076}{{\tt arXiv:1912.07076}}.

\end{thebibliography}




\end{document}